\documentclass[10pt,twocolumn,letterpaper]{article}

\usepackage{cvpr}
\usepackage{times}
\usepackage{epsfig}
\usepackage{graphicx}
\usepackage{amsmath}
\usepackage{amssymb}
\usepackage{url}
\usepackage{multirow}
\usepackage[noend]{algpseudocode}
\usepackage{algorithmicx,algorithm}
\usepackage{bm}
\usepackage{dsfont}
\usepackage{subfig}
\usepackage{epstopdf}
\usepackage[T1]{fontenc}
\usepackage[utf8]{inputenc}
\usepackage{authblk}
\usepackage{color}
\usepackage[font=footnotesize,labelfont=bf]{caption}
\usepackage{booktabs}

\newcommand{\tabincell}[2]{\begin{tabular}{@{}#1@{}}#2\end{tabular}}

\cvprfinalcopy 

\def\cvprPaperID{1706} 

\ifcvprfinal\fi

\begin{document}

\title{LG-GAN: Label Guided Adversarial Network for Flexible Targeted Attack of Point Cloud-based Deep Networks}

\author{Hang Zhou$^{1}$\thanks{Equal contribution, $\dagger$ Corresponding author}, Dongdong Chen$^{2,*}$, Jing Liao$^{3}$,  Kejiang Chen$^{1}$ , Xiaoyi Dong$^{1}$, \\ Kunlin Liu$^{1}$, Weiming Zhang$^{1,\dagger}$, Gang Hua$^{4}$, Nenghai Yu$^{1}$\\
$^{1}$University of Science and Technology of China, \quad $^{2}$Microsoft Research \\
$^{3}$The Chinese University of Hong Kong, $^{4}$Wormpex AI Research
\\
{\tt\small\{zh2991,chenkj,dlight,lkl6949\}@mail.ustc.edu.cn};
{\tt\small cddlyf@gmail.com};\\
{\tt\small jingliao@cityu.edu.hk}; 
{\tt\small\{zhangwm, ynh\}@ustc.edu.cn};
{\tt\small ganghua@gmail.com}
}

\maketitle
\ifcvprfinal\thispagestyle{empty}\fi

\begin{abstract}
Deep neural networks have made tremendous progress in  3D point-cloud recognition. Recent works have shown that these 3D recognition networks are also vulnerable to adversarial samples produced from various attack methods, including optimization-based 3D Carlini-Wagner attack, gradient-based iterative fast gradient method, and skeleton-detach based point-dropping. However, after a careful analysis, these methods are either extremely slow because of the optimization/iterative scheme, or not flexible to support targeted attack of a specific category. To overcome these shortcomings, this paper proposes a novel label guided adversarial network (LG-GAN) for real-time flexible targeted point cloud attack. To the best of our knowledge, this is the first generation based 3D point cloud attack method. By feeding the original point clouds and target attack label into LG-GAN, it can learn how to deform the point clouds to mislead the recognition network into the specific label only with a single forward pass. In detail, LG-GAN first leverages one multi-branch adversarial network to extract hierarchical features of the input point clouds, then incorporates the specified label information into multiple intermediate features using the label encoder. Finally, the encoded features will be fed into the coordinate reconstruction decoder to generate the target adversarial sample. By evaluating different point-cloud recognition models (\eg, PointNet, PointNet++ and DGCNN), we demonstrate that the proposed LG-GAN can support flexible targeted attack on the fly while guaranteeing good attack performance and higher efficiency simultaneously.
\end{abstract}

\section{Introduction}
Deep neural networks (DNNs) have been successfully applied to many computer vision tasks~\cite{szegedy2015going,he2016deep,ren2015faster,chen2017coherent,zhang2020model,mne_iccv19}. Recently, pioneering works such as DeepSets~\cite{zaheer2017deep}, PointNet~\cite{charles2017pointnet} and its variants~\cite{qi2017pointnet++,wang2019dynamic} explored the possibility of reasoning with point clouds through DNNs for understanding geometry and recognizing 3D structures. These DNN-based methods directly extract features from raw 3D point coordinates without utilizing hand-crafted features, such as normals and curvatures, and present impressive results for 3D object classification and semantic scene segmentation.

However, recent works have demonstrated that DNNs are vulnerable to adversarial examples, which are maliciously created by adding imperceptible perturbations to the original input by attackers. This would potentially bring security threats to real application systems such as autonomous driving~\cite{cao2019adversarial}, speech recognition~\cite{cisse2017houdini} and face verification~\cite{liu2016delving}, to name a few. Similar to images, many recent works ~\cite{xiang2019generating,liu2019extending,zheng2019pointcloud} have shown that deep point cloud recognition networks are also sensitive to adversarial examples and readily fooled by them.

Existing 3D adversarial attack methods can be roughly categorized into three: optimization-based methods such as L-BFGS~\cite{szegedy2013intriguing} and C\&W attack~\cite{carlini2017towards}, gradient-based methods such as FGSM~\cite{goodfellow2014explaining} and IFGM~\cite{gu2014towards}, and skeleton-detach based methods like \cite{yang2019adversarial}. Notwithstanding their demonstrated effectiveness in attack, they all have some significant limitations. For example, the former two rely on a very time-consuming optimization/iterative scheme. This renders them not flexible enough for real-time attack. In contrast, the last one is faster but has lower attack success rates. Moreover, it requires to delete some critical point cloud structures and only supports the untargeted attack.

To overcome these shortcomings and motivated by image adversarial attack methods~\cite{baluja2017adversarial,xiao2018generating,han2019once}, this paper proposes the first generation based 3D point cloud attack method, which is faster, has better attack performance and supports flexible targeted attack  of a specific category on the fly.

Specifically, we design a novel label guided adversarial network ``LG-GAN''. By feeding the original point clouds and target attack label into LG-GAN, it can learn how to deform the point cloud with minimal perturbations to mislead the recognition network into the specific label only with a single forward pass. Specifically, LG-GAN first leverages one multi-branch generative network to extract hierarchical features of the input point clouds, then incorporates the specified label information
into multiple intermediate features by a label encoder. 
Finally, the transformed features will be fed into the coordinate reconstruction decoder to generate the target adversarial sample. To further encourage the ``LG-GAN'' to produce a visually pleasant point cloud, a graph patch-based discriminator network is leveraged for adversarial training.

To demonstrate the effectiveness, we evaluate the proposed ``LG-GAN'' by attacking different state-of-the-art point-cloud recognition models (\eg, PointNet, PointNet++ and DGCNN). Experiments demonstrate that it can achieve good attack performance and better efficiency simultaneously for flexible targeted attack.

In summary, our contributions are three fold:
\begin{itemize}
\item Motivated by the limitations of existing 3D adversarial attack methods, we propose the first generation based adversarial attack method for 3D point-cloud recognition networks.

\item To support arbitrary-target attack, we design a novel label guided adversarial network ``LG-GAN'' by multiple intermediate feature incorporation.

\item Experiments on different recognition models demonstrate that our method is both more flexible and effective in targeted attack while being more efficient.

\end{itemize}

\section{Related Work}
\begin{figure*}
\begin{center}
{\centering\includegraphics[width=1.0\linewidth]{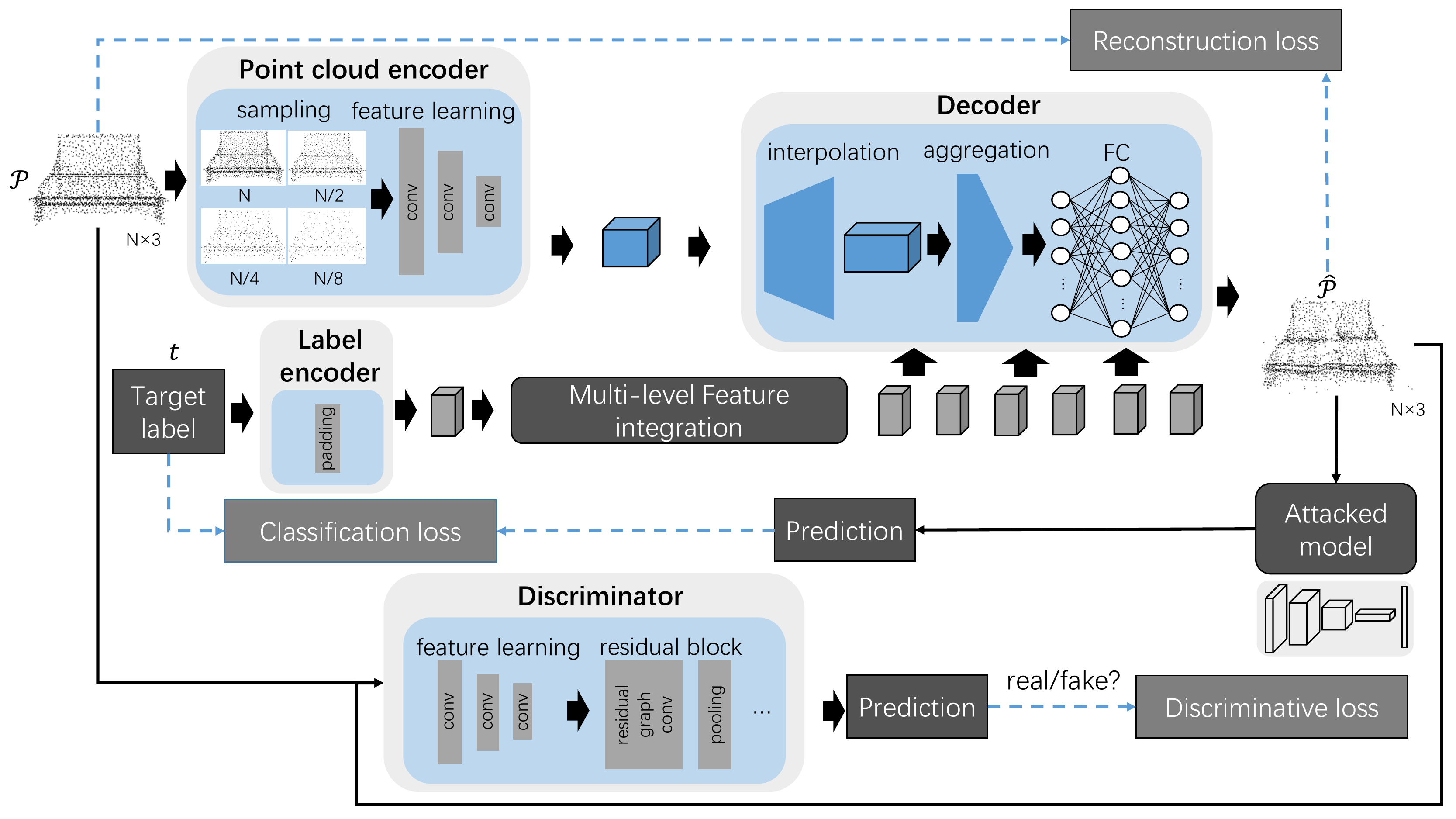}
}\\
\caption{\textbf{Illustration of the LG-GAN architecture} for training the generation of targeted 3D adversarial point clouds. Given an input point cloud $\mathcal{P}$ of $N$ points with XYZ coordinates, a hierarchical feature embedding network (implemented with PointNet++~\cite{qi2017pointnet++} layers) subsamples and learns deep features on the points. Besides, a label encoder transforms the label information into a label feature to be better concatenated with the point cloud feature at each feature level. The decoder network includes interpolation, aggregation, and FC layers. The discriminator network $\mathcal{D}$ includes a feature learning net and multiple residual blocks. Image best viewed in color. }\label{fig:01}
\end{center}
\end{figure*}

\subsection{3D point recognition}
To obtain great 3D point recognition performance, the most important component is the feature extractor design. In the era before deep learning, various kinds of handcraft 3D descriptors are proposed. For example, the shape distribution is exploited in \cite{osada2002shape} to calculate the similarity based on distance, angle, area, and volume between random surface points. Based on the native 3D representations of objects, classical shape-based descriptors include voxel grid \cite{osada2002shape}, polygon mesh \cite{kokkinos2012intrinsic}, and 3D SIFT and SURF descriptors \cite{scovanner20073,knopp2010hough}. Recently, thanks to the strong learning capability of deep networks, more deep learning-based descriptors are proposed.

In \cite{su2015multi}, a multi-view convolutional neural network (MVCNN) is proposed to fuse multiple-view features by a pooling procedure. The pioneering work PointNet \cite{qi2017pointnet++} further designs a novel neural network that can directly consume point clouds and achieves superior recognition performance. By considering more local structure information, it is further improved in the following work PointNet++ \cite{qi2017pointnet++}. Another representative work is DGCNN \cite{wang2019dynamic}, which exploits local geometric structures by constructing a
local neighborhood graph and applying convolution-like operations on the edges connecting neighboring pairs of points in the spirit of graph neural networks. To evaluate the effectiveness and generalization ability of our method, we will try to attack these three representative methods respectively.

\subsection{Existing 3D adversarial attack methods}
As discussed before, 3D recognition networks are also vulnerable to adversarial attacks. We can roughly divide existing 3D adversarial attack methods into three different categories: optimization-based methods \cite{xiang2019generating}, gradient-based methods \cite{liu2019extending,yang2019adversarial,zheng2019pointcloud}, and skeleton-based methods \cite{yang2019adversarial}. For optimization-based methods, Xiang \etal~\cite{xiang2019generating} propose a C\&W  based framework~\cite{carlini2017towards} to generate adversarial point clouds by point perturbation, point adding and cluster adding. This method uses an optimization objective to seek the minimal perturbated sample that can make the recognition network classify incorrectly.

Typical perturbation measurements include $L_2$ norm, Hausdorff distance, and Chamfer distance. For gradient-based methods, Liu \etal~\cite{liu2019extending} extended the fast/iterative gradient method by constraining the perturbation magnitude onto an epsilon ball surface in different dimensions. Yang \etal~\cite{yang2019adversarial} developed a variant of one-pixel attack~\cite{su2019one} by using pointwise gradient method to only update the attached points without changing the original point cloud. Zheng \etal~\cite{zheng2019pointcloud} proposed point dropping based attack by first constructing a gradient-based salience map and dropping points with lowest salience scores.

Though optimization-based and gradient-based methods can achieve pretty good attack performance, they are both extremely slow. Motivated by the fact that the recognition result of one 3D object in PointNet ~\cite{qi2017pointnet} is often determined by a critical subset, Yang \etal~\cite{yang2019adversarial} developed a skeleton-detach based attack method by iteratively detaching the most important points from this critical subset. This method is faster but its attack success rate is not that high. More importantly, this method cannot support targeted attack. To overcome all the aforementioned limitations, we propose the first generation based attack method, which enables faster and better flexible targeted attacking.

\subsection{Existing 3D adversarial defense methods}
To defend against the 3D adversarial attacks, we can derive from the well-known image adversarial defense strategies. For example, by augmenting the training set with adversarial examples, adversarial training~\cite{goodfellow2014explaining,kurakin2018ensemble} can significantly increase the model's robustness. Other simple defense methods include random point sampling, Gaussian noising and quantification. Recently, Zhou~\etal~\cite{zhou2019dup} deploy a statistical outlier denoiser and a data-driven upsampling network as the pre-processing operation before the input is fed into the recognition network. The denoiser contributes to removing outlier based noise patterns as a non-differentiable layer while the upsampling network help purify the perturbed point clouds. In this paper, these defense methods are considered to evaluate the attack performance.


\section{Generation-based 3D Point Cloud Attack}
Motivated by the shortcomings of existing methods, we propose a novel generation based 3D point cloud attack method ``LG-GAN'', as shown in Fig.~\ref{fig:01}. It consists of two sub-networks: a generative network $\mathcal{G}$ that learns how to transform the input point clouds into targeted adversarial samples and a discriminator network $\mathcal{D}$ that encourages the outputs of $\mathcal{G}$ indistinguishable from clean point clouds. These two sub-networks are trained in an adversarial manner, and the discriminator $\mathcal{D}$ is just an auxiliary network and not needed anymore after training.

\subsection{Label guided Adversarial network $\mathcal{G}$}
Given a clean point cloud $\mathcal{P}$ of total $N$ points and the target label $t$ to which we want the recognition network to misclassify, $\mathcal{G}$ aims to learn how to transform $\mathcal{P}$ into an adversarial sample $\mathcal{\hat{P}}$ with minimal perturbations to $\mathcal{P}$ based on $t$, which is constrained by three designed losses (classification loss, reconstruction loss and discriminative loss). It mainly consists of three different parts: a label encoder $\mathbf{E}_l$, a hierarchical point feature encoder $\mathbf{E}_p$, and a point decoder $\mathbf{D}_p$. Specifically, $\mathbf{E}_l$ is responsible for encoding label $t$ into a latent label code $z_t$ by repeating $t$ for multiple times to match the point number of each feature layer $i$, and $\mathbf{E}_p$ encodes $\mathcal{P}$ into a multi-level feature embedding $\overrightarrow{F_\mathcal{P}} = (F_\mathcal{P}^1, ..., F_\mathcal{P}^m)$, then $z_t$ will be concatenated with $\overrightarrow{F_\mathcal{P}}$ and fed into $\mathbf{D}_p$ to obtain the final adversarial sample $\mathcal{P}_{adv}$. Mathematically:

\begin{equation}
\begin{aligned}
    \overrightarrow{z_t} &= (z_t^1,...,z_t^l)=\mathbf{E}_l(t), \\
    \overrightarrow{F_\mathcal{P}} &= (F_\mathcal{P}^1, ..., F_\mathcal{P}^m) = \mathbf{E}_p (\mathcal{P}), \\
    \mathcal{P}_{adv} &= \mathbf{D}_p(\overrightarrow{z_t}, \overrightarrow{F_\mathcal{P}}).
\end{aligned}
\end{equation}
Here $m$ is the feature level number ($4$ by default). Note that $t$ is formatted as a one-hot vector whose $t$th value is 1.

%

\vspace{1em}
\noindent \textbf{Hierarchical Point Feature Learning.} Progressively extracting features of different scales in a hierarchical manner has been proven to be an effective strategy for capturing both the local and global point structures. By default, in this paper, we use $4$ different levels of point cloud representation from coarse to fine. Specifically, given the point cloud $\mathcal{P}$ consisting of $N$ points, we use the iterative farthest point sampling algorithm as the sampling layer to evenly sample the points into four scales $(N, N/2, N/4, N/8)$. Then we adopt PointNet++ \cite{qi2017pointnet++} to extract the feature embedding for each scale. In detail, for each point in level $i$, PointNet++ will utilize the local structure information and aggregate the features of neighboring points as the final feature $F_P^i$, whose shape is $\frac{N}{2^{i-1}}\times (64\cdot 2^{i-1})$.



\vspace{1em}
\noindent\textbf{Feature Decoding and Label Concatenation to Reconstruct the Final Adversarial Sample.} To aggregate the multi-scale features from each level, we follow the strategy used in \cite{hou2017deeply,hariharan2015hypercolumns,yu2018pu} that directly combine features from different levels and let the network learn the importance of each level. Since the different scales have different point number, the downsampled point features will be upsampled to the original point number by using the interpolation method adopted in PointNet++ ~\cite{qi2017pointnet++}. Formally, denote the upsampled feature as ${F_\mathcal{P}^i}'$, then:
\begin{equation}
\label{eqn:1b}
{F^i_\mathcal{P}}'(x)=\frac{\sum^{3}_{j=1}w_j(x)F^i_\mathcal{P}(x_j)}{\sum^{3}_{j=1}w_j(x)},
\end{equation}
where $w_j(x)$ is the contribution weight of the neighborhood point $x_j$ defined as $1/d(x,x_j)$, and $d$ is the $\ell_2$ distance by default. Then each scale upsampled features will be processed by one $1\times 1$ convolutional layer similar to that in \cite{yu2018pu} to have the same dimension.

The $m$ level of features ${F^i_\mathcal{P}}'(x)$ together with the original point cloud coordinates are concatenated together along the feature dimension to obtain the final aggregated feature ${F_\mathcal{P}}''(x)$. Then, $l$ fully connected layers with multi-layered label concatenation are utilized to reconstruct the detailed coordinates of the adversarial sample. At each fully connected layer $i$ (except for the last layer), we obtain the integrated feature:
\begin{equation}
{F_\mathcal{P}^{i+1}}''(x) = \textrm{FC}([z_t^i, {F_\mathcal{P}^i}''(x)]), i=1,...,l-1.
\end{equation}
Here $[x,y]$ means concatenating $x, y$ along the feature dimension and $l$ is the number of fully connected layer ($4$ by default).


\subsection{Graph Discriminator $\mathcal{D}$}
To help generate a more realistic adversarial point cloud, we further leverage a graph discriminator network $\mathcal{D}$ for adversarial learning. For the detailed network structure, we directly use the existing graph patch GAN~\cite{shrivastava2017learning,wu2019point} to distinguish clean point clouds from generated adversarial point clouds. Then for every locally generated patch, $\mathcal{D}$ will encourage it to lie on the distribution of the clean point clouds.

\begin{figure}
\begin{center}
{\centering\includegraphics[width=0.80\linewidth]{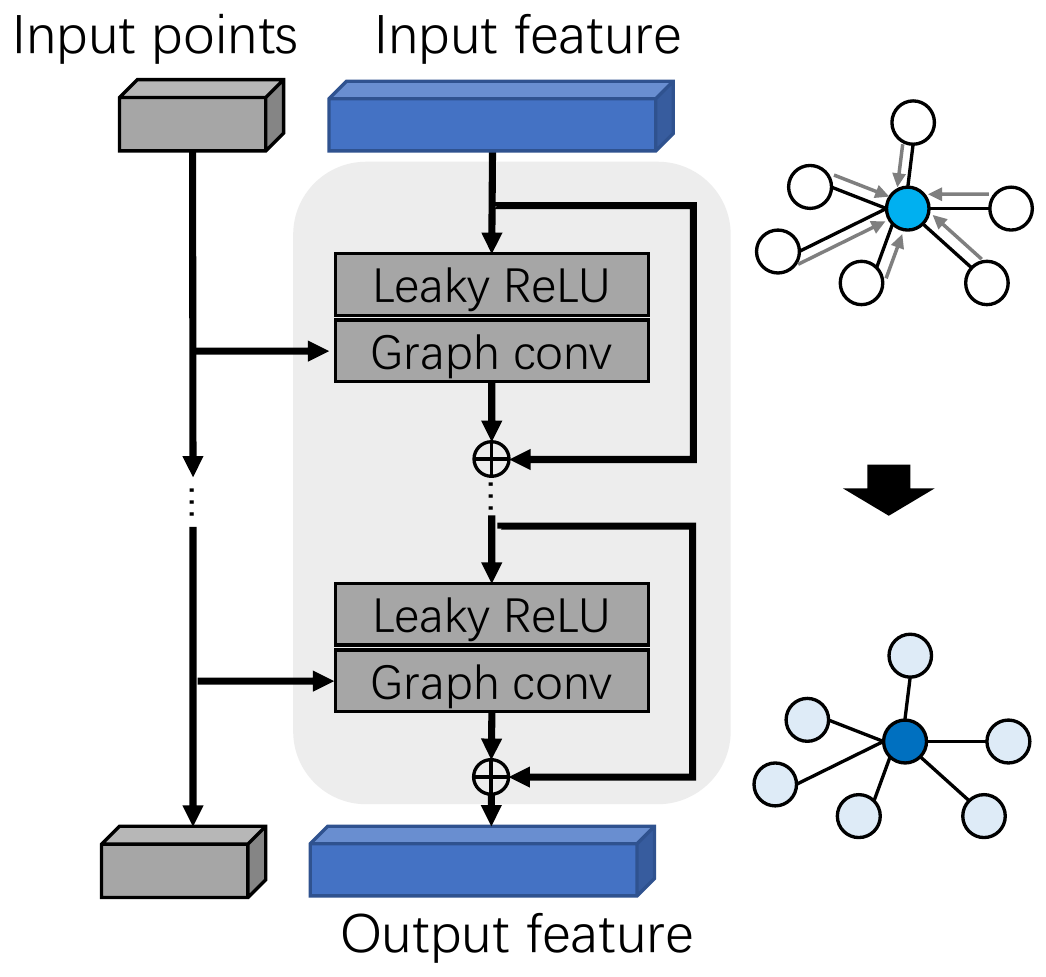}
}\\
\caption{\textbf{Residual graph convolutional block}. The input points are utilized for querying the $k$ nearest neighbors. The output feature of the block has the same dimension as the input feature. }\label{fg:gcnn}
\end{center}
\end{figure}

For the detailed network structure, $\mathcal{D}$ consists of a PointNet++ like feature extraction head and a series of pooling blocks and residual graph convolution blocks. For the feature extraction head, it consists of several pointwise convolutional layers and aggregate the features of $k$ nearest neighbors for each point by max pooling. For the pool block, it will first use the farthest point sampling algorithm to find some seed points then max-pool the corresponding features of their neighboring points to get the downsampled feature. For the residual graph convolution block, it is composed of several graph convolutions defined on the $k$ nearest neighborhood-based graph $G=(v,N(x))$ and residual connections as shown in Fig.~\ref{fg:gcnn}. Mathematically, the formulation of graph convolution is defined as:
\begin{equation}
\label{eqn:1t}
f_{out}(x)=w_0f_{in}(x)+w_1\sum_{q\in N(x)}f_{in}(q), \quad\forall x\in v,
\end{equation}
where $N(x)$ is defined by the $k$ nearest neighbors of $x$ in the Euclidean space. $f_{in}, f_{out}$ are the input and output features of the vertex $x$ respectively. $w_0,w_1$ are the learnable weights that determine how much information needs to be borrowed from the target vertex and its neighborhood vertexes. $k$ is set to 8 by default.


\subsection{Objective Loss Function}
The objective loss function of the generator networks $\mathcal{G}$ consists of three parts: the classification loss, the reconstruction loss, and the discriminative loss. Formally:
\begin{equation}
\label{eqn:1t}
\mathcal{L}_{\mathcal{G}}=\mathcal{L}_{cls}+\alpha\mathcal{L}_{rec}+\beta\mathcal{L}_{dis},
\end{equation}
where $\alpha$, $\beta$ are weight factors.
$\mathcal{L}_{cls}$ is the classification loss that urges the attacked model $\mathcal{H}$ to make prediction to the target label $t$, and we use the standard cross-entropy loss:
\begin{equation}
\label{eqn:1t}
\mathcal{L}_{cls}=-[t\log{\mathcal{H}(\hat{\mathcal{P}})}+(1-t)\log{(1-\mathcal{H}(\hat{\mathcal{P}}))}],
\end{equation}
where $\hat{\mathcal{P}}=\mathcal{G}_{\theta}(\mathcal{P},t)$ is the generated adversarial point cloud, $\mathcal{H}(\hat{\mathcal{P}})$ is the predicted probability of the target model on the adversarial sample. $\mathcal{L}_{rec}$ is the reconstruction loss that encourages the generated adversarial point cloud to resemble the original sample. We adopt an $\ell_2$ distance, which is better than the Chamfer distance, as our measurement.
$\mathcal{L}_{dis}$ is a graph adversarial loss that has the same goal with $\mathcal{L}_{rec}$. Inspired by LS-GAN~\cite{mao2017least},
\begin{equation}
\label{eqn:1w}
\mathcal{L}_{dis}(\hat{\mathcal{P}})=\lVert 1-\mathcal{D}_{\theta}(\hat{\mathcal{P}}) \rVert_2^2.
\end{equation}

The objective loss function of the discriminator networks $\mathcal{D}$ aims to distinguish real and fake point clouds by minimizing $\mathcal{L}_{\mathcal{D}}$ loss:
\begin{equation}
\label{eqn:1v}
\mathcal{L}_{\mathcal{D}}(\mathcal{P},\hat{\mathcal{P}})=\frac{1}{2}\lVert \mathcal{D}_{\theta}(\hat{\mathcal{P}}) \rVert_2^2+\frac{1}{2}\lVert 1-\mathcal{D}_{\theta}(\mathcal{P}) \rVert_2^2.
\end{equation}

\subsection{Implementation Details}
Following the practice in~\cite{carlini2017towards}, we adopt PointNet~\cite{qi2017pointnet} and PointNet++~\cite{qi2017pointnet++} as the targeted attack models $\mathcal{H}$, and use the default settings to train.
Given the pretrained models, we train the proposed LG-GAN to attack the two models. The size of input point cloud is $2,048\times 3$, and weights $\alpha$ and $\beta$ are set with 0.001 and 1 respectively. The implementation is based on TensorFlow. For the optimization, we train the network for 200 epochs using the Adam~\cite{kingma2014adam} optimizer with a minibatch size of 4, and the learning rate of $\mathcal{G}$ and $\mathcal{D}$ are 0.001 and 0.00001 respectively. The whole training process takes about 8h on the NVIDIA RTX 2080 Ti GPU.

\section{Experiments}

In this section, we firstly compare our LG-GAN with previous state-of-the-art methods on a CAD object benchmark (Sec. \ref{compare}). We then provide analysis experiments to understand the effectiveness and efficiency of LG-GAN (Sec. \ref{ablation}). We also validate translation-based attack (Sec. \ref{translation}). Finally we show qualitative results of our LG-GAN (Sec. \ref{qualitative}). More analysis and visualizations are provided in the supplementary materials.

\begin{table*}
\footnotesize
\begin{center}
\begin{tabular}{c|c|c|c|c|c|c}
\hline\toprule[0.4pt]
 & Target~\cite{charles2017pointnet} & \tabincell{c}{Defense\\(SRS)~\cite{zhou2019dup}} & \tabincell{c}{Defense\\ (DUP-Net)~\cite{zhou2019dup}} & \tabincell{c}{$\ell_2$ dist\\(meter)} & \tabincell{c}{Chamfer dist\\(meter)} & \tabincell{c}{Time\\(second)}\\
\midrule
C\&W $+$ $\ell_2$~\cite{xiang2019generating}    & 100  & 0    & 0   & \textbf{0.01} & 0.006 & 40.80 \\
C\&W $+$ Hausdorff~\cite{xiang2019generating}   & 100  & 0    & 0   & ---    & 0.005 & 42.67 \\
C\&W $+$ Chamfer~\cite{xiang2019generating}     & 100  & 0    & 0   & ---    & \textbf{0.005} & 43.73 \\
C\&W $+$ 3 clusters~\cite{xiang2019generating}  & 94.7 & 2.7  & 0   & ---    & 0.120 & 52.00 \\
C\&W $+$ 3 objects~\cite{xiang2019generating}   & 97.3 & 3.1  & 0   & ---    & 0.064 & 58.93 \\
FGSM~\cite{liu2019extending,yang2019adversarial}& 12.2 & 5.2  & 2.8 & 0.15   & 0.129 & 0.082 \\
IFGM~\cite{liu2019extending,yang2019adversarial}& 73.0 & 14.5 & 3.3 & 0.31   & 0.132 & 0.275 \\
\midrule
LG $+$ Chamfer (ours)                    & 96.1 & 75.4 & 13.9 & 0.63 & 0.137 & 0.037 \\
single-layered LG-GAN (ours)             & 97.6 & 80.2 & 37.8 & 0.27 & 0.032 & 0.053 \\
LG (ours)                                & 97.1 & 85.0 & 72.0 & 0.25 & 0.028 & \textbf{0.033} \\
LG-GAN (ours)                            & \textbf{98.3} & \textbf{88.8} & \textbf{84.8} & 0.35 & 0.038 & 0.040 \\
\hline\toprule[0.4pt]
\end{tabular}
\end{center}
\caption{\textbf{Attack success rate (\%, second to fourth column), distance (fifth-sixth column) between original sample and adversarial sample (meter per object) and generating time (second per object) on attacking PointNet from ModelNet40.} ``Target'' stands for white-box attacks. The hyper-parameter setting of two gray-box attacks is: for the simple random sampling (SRS) defense model, percentage of random dropped points is 60\%$\sim$90\%; for DUP-Net defense model, $k=50$ and $\alpha=0.9$ from~\cite{zhou2019dup}. The default LG-GAN (ours) consists of multi-layered label embedding, $\ell_2$ loss and GAN loss. }
\label{tab:01}
\end{table*}

\subsection{Comparing with State-of-the-art Methods}\label{compare}
\textbf{Dataset. }
ModelNet40~\cite{wu20153d} is a comprehensive clean collection of 3D CAD models for objects, which contains 12,311 objects from 40 categories, where 9,843 are used for training and the other 2,468 for testing. As done by Qi \etal~\cite{qi2017pointnet}, before generating adversarial point clouds, we first uniformly sample 2,048 points from the surface of each object and rescale them into a unit cube.

ShapeNetCore~\cite{chang2015shapenet} is a subset of the full ShapeNet dataset with single clean 3D models and manually verified category and alignment annotations. It covers 55 common object categories with 52,472 unique 3D models, where 41,986 are used for training and the other 10,486 for testing. All the data are uniformly sampled into 4096 points.

\vspace{0.5em}
\textbf{Attack evaluations. }
The attackers generate adversarial examples using the targeted models and then evaluate the attack success rate and detection accuracy of these generated adversarial examples on the target and defense models.

\vspace{0.5em}
\textbf{Methods in comparison. }
We compare our method with a wide range of prior art methods. C\&W~\cite{xiang2019generating} is an optimization based algorithm with various loss criteria including $\ell_2$ norm, Hausdorff and Chamfer distances, cluster and object constraints. FGSM and IFGM~\cite{liu2019extending,yang2019adversarial} are gradient-based algorithms constrained by the $\ell_{\infty}$ and $\ell_2$ norms, where FGSM is a na\"ive baseline that subtracts perturbations along with the direction of the sign of the loss gradients with respect to the input point cloud, and IFGM iteratively subtracts the $\ell_2$-normalized gradients. Since FGSM is not strong enough to handle targeted attack (less than 30\% attack success rate) and point-detach methods~\cite{qi2017pointnet,yang2019adversarial} cannot targetedly attack pretrained networks, we do not compare these methods in the following experiments.

\vspace{0.5em}
\textbf{Results} are summarized in Table~\ref{tab:01} and Table~\ref{tab:02}. FGSM has low attack success rates (12.2\%). LG-GAN outperforms IFGM methods by at least 23.1\% of attack success rates and $5\times$ of generation speed. LG-GAN has similar attack success rate with optimization-based C\&W methods, but outperforms speed by $100\times$. In terms of visual quality, LG-GAN performs a little worse than C\&W methods since C\&W attempt to modify points as little as possible to implement attacks by querying the network multiple times, and thus it is difficult to attack in real time. LG-GAN has better attack ability on robust defense models (gray-box models), with 88.8\% and 84.8\% attack success rates on simple random sampling (SRS) and DUP-Net~\cite{zhou2019dup} defense models, respectively. Notably, we achieve great improvements with regard to attack black-box models. As shown in Table~\ref{tab:02}, LG-GAN has 11.6\% and 14.5\% attack success rates on PointNet++~\cite{qi2017pointnet++} and DGCNN~\cite{wang2019dynamic} respectively while IFGM~\cite{liu2019extending,yang2019adversarial} only has 3.0\%, 2.6\% and C\&W~\cite{xiang2019generating} based methods with 0 and 0.

\begin{table*}
\footnotesize
\begin{center}
\begin{tabular}{c|c|c|c|c|c|c|c|c}
\hline\toprule[0.4pt]
 & Attack type & \tabincell{c}{C\&W $+$\\$\ell_2$~\cite{xiang2019generating}} & \tabincell{c}{C\&W $+$\\Chamfer~\cite{xiang2019generating}} & \tabincell{c}{IFGM~\\\cite{liu2019extending,yang2019adversarial}} & \tabincell{c}{LG $+$\\Chamfer (ours)} & \tabincell{c}{single-layered\\LG-GAN (ours)} & LG (ours) & \tabincell{c}{LG-GAN\\(ours)} \\
\midrule
 \textbf{PointNet}~\cite{qi2017pointnet}    & White-box & 100/0 & 100/0  & 73.0/0.4 & 96.1/0.45 & 97.6/0.16 & 97.6/0.45 & \textbf{98.3}/0.54 \\
\midrule
\textbf{PointNet++}~\cite{qi2017pointnet++} & Black-box & 0/3.1 & 0/6.2  & 3.0/3.4  & 2.8/4.1   & 9.4/5.3   & 6.1/8.0    & \textbf{11.6}/5.2 \\
\midrule
\textbf{DGCNN}~\cite{wang2019dynamic}       & Black-box & 0/9.8 & 0/5.3  & 2.6/4.1  & 5.8/6.2   & 13.5/7.4  & 9.4/8.2    & \textbf{14.5}/15.3\\
\hline\toprule[0.4pt]
\end{tabular}
\end{center}
\caption{\textbf{Black-box attack success rate/accuracy (\%).} The adversarial point clouds of ModelNet40 are generated from PointNet~\cite{qi2017pointnet}, and then attack PointNet++~\cite{qi2017pointnet++} and DGCNN~\cite{hua2018pointwise}. }
\label{tab:02}
\end{table*}

\begin{figure}
\begin{center}
\centering
\includegraphics[width=0.95\linewidth]{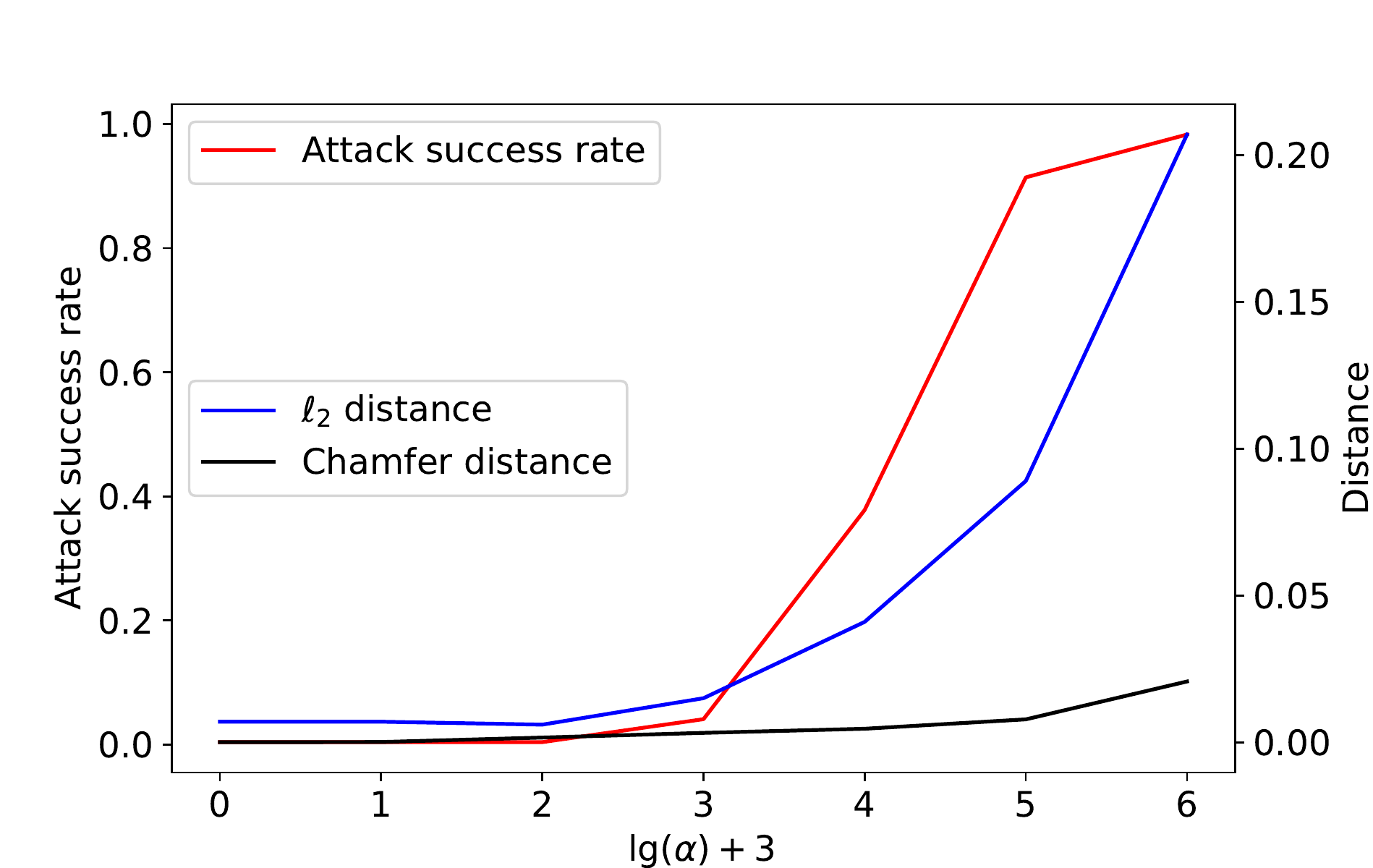}
\\
\caption{\textbf{Larger $\alpha$ helps more generated point clouds successfully target-attack PointNet, but affects visual quality of generated objects from ModelNet40.} We show the average attack success rate (in red lines) of LG-GAN \wrt balancing weight $\alpha$, (in blue lines) the average $\ell_2$ distance, and (in black lines) the average Chamfer distance.}
\label{fig:alpha}
\end{center}
\end{figure}

\begin{figure}
\begin{center}
\centering
\includegraphics[width=1.00\linewidth]{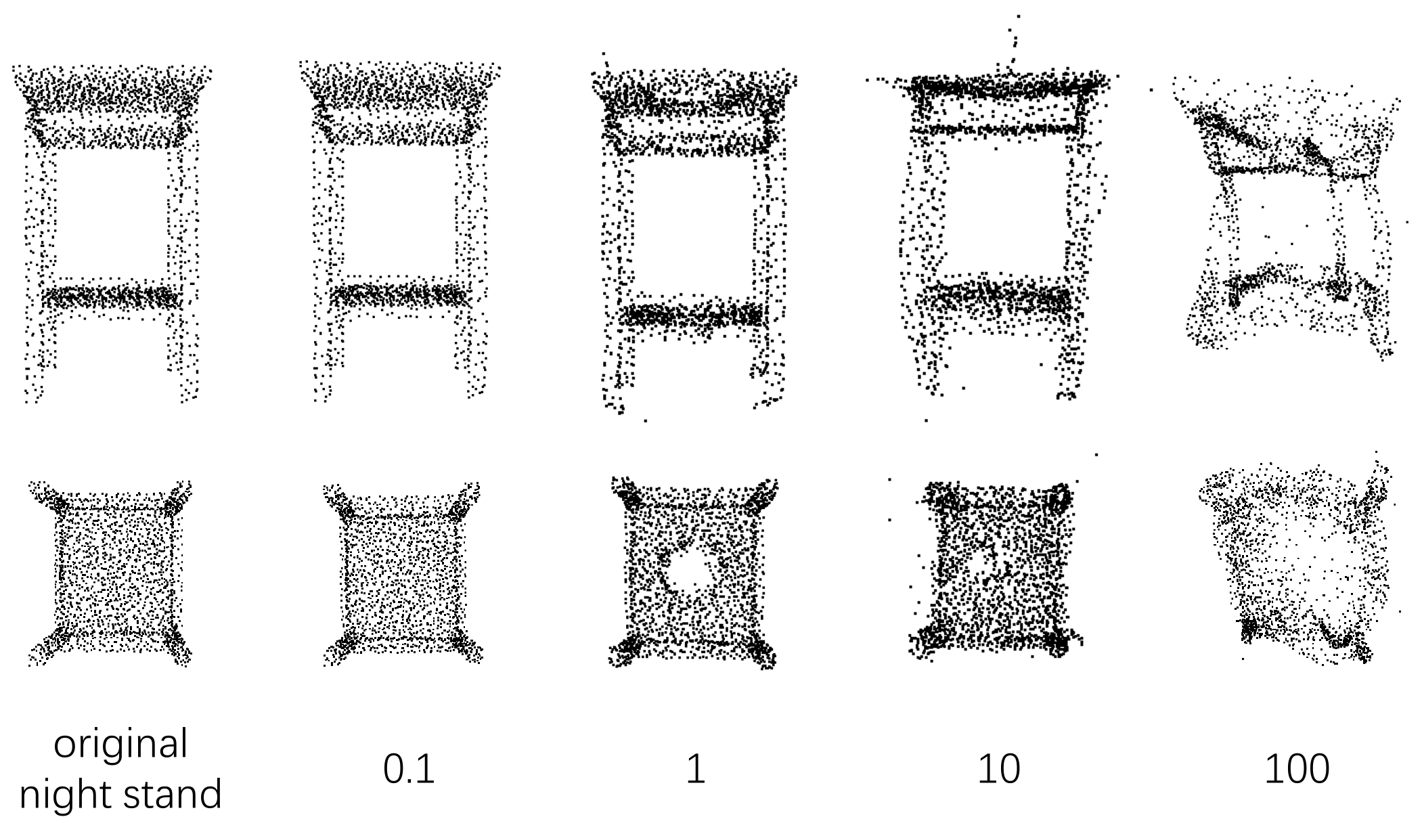}
\\
\caption{\textbf{Qualitative results of object ``night stand'' with different $\alpha$.} It shows (from left to right): the original point cloud from ModelNet40, the generated point cloud with $\alpha$ ranging among 0.1, 1, 10, and 100. Images from top to bottom are the front view and bottom view. }
\label{fig:alpha_img}
\end{center}
\end{figure}

\begin{figure*}
\begin{center}
\centering
\subfloat[]{
	\label{fig:subfig_a}
	\includegraphics[width=0.63\linewidth]{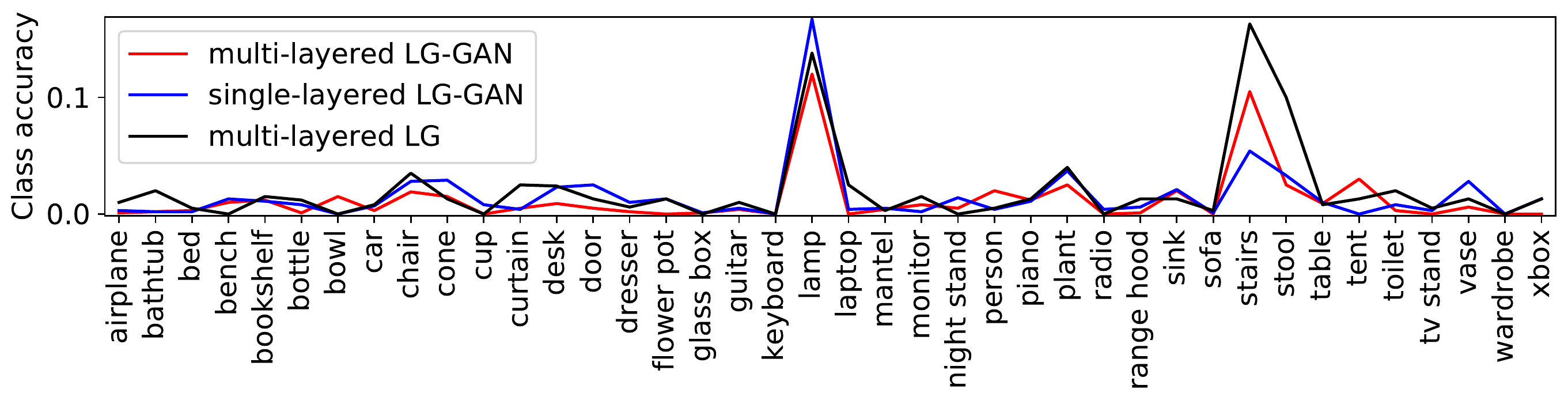}
}
\centering
\subfloat[]{
	\label{fig:subfig_b}
	\includegraphics[width=0.35\linewidth]{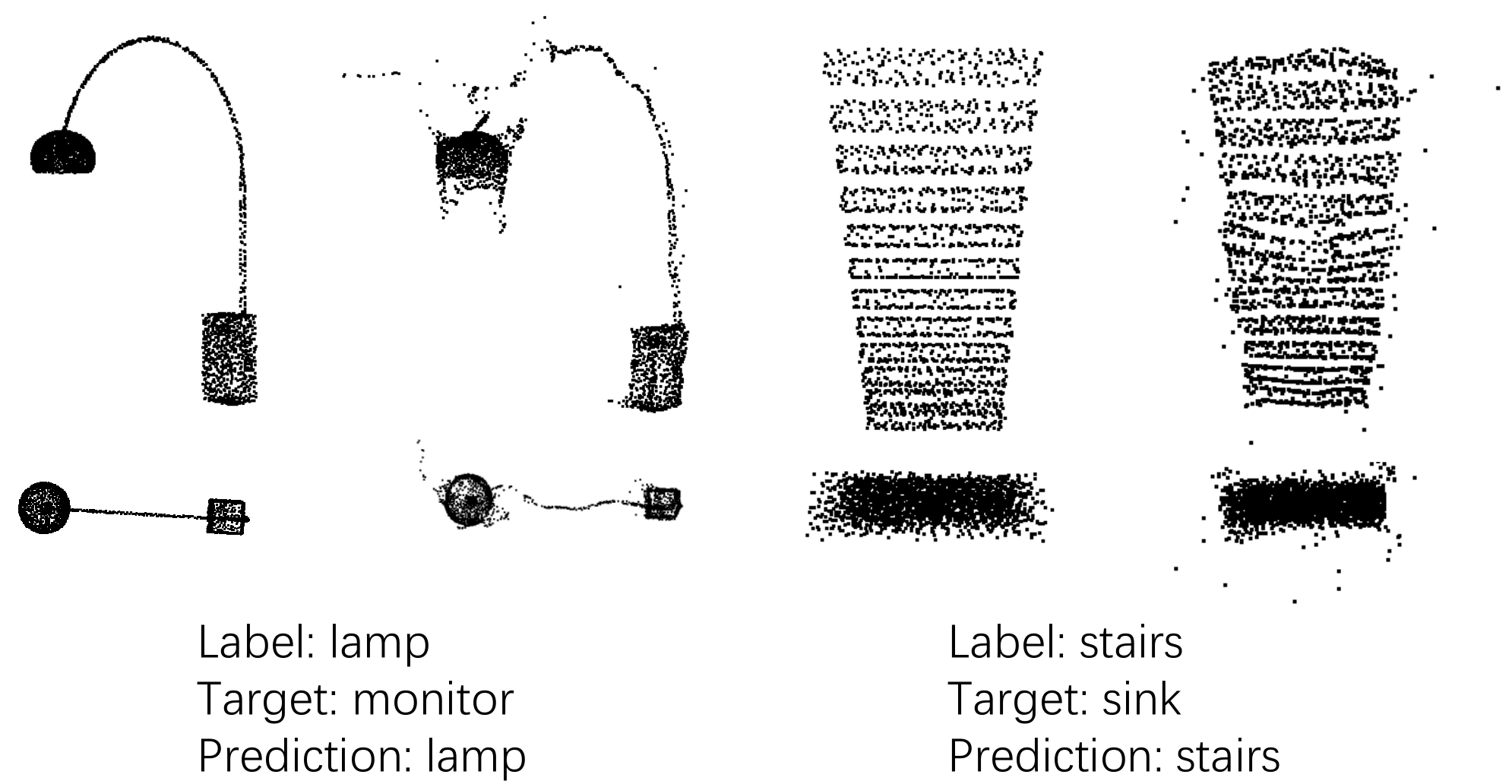}
}\\
\caption{\textbf{(a) Class accuracy of generated adversarial point clouds of ModelNet40} averaged over 10 random testing. Some objects from ``lamp'' and ``stairs'' classes are harder to targetedly attack; \textbf{(b) Visualization of difficult crafts} with failed attacks: objects from ``lamp'' and ``stairs'' categories and their adversarial examples. }
\label{fig:class_acc}
\end{center}
\end{figure*}


\begin{figure}
\begin{center}
{\centering\includegraphics[width=1.00\linewidth]{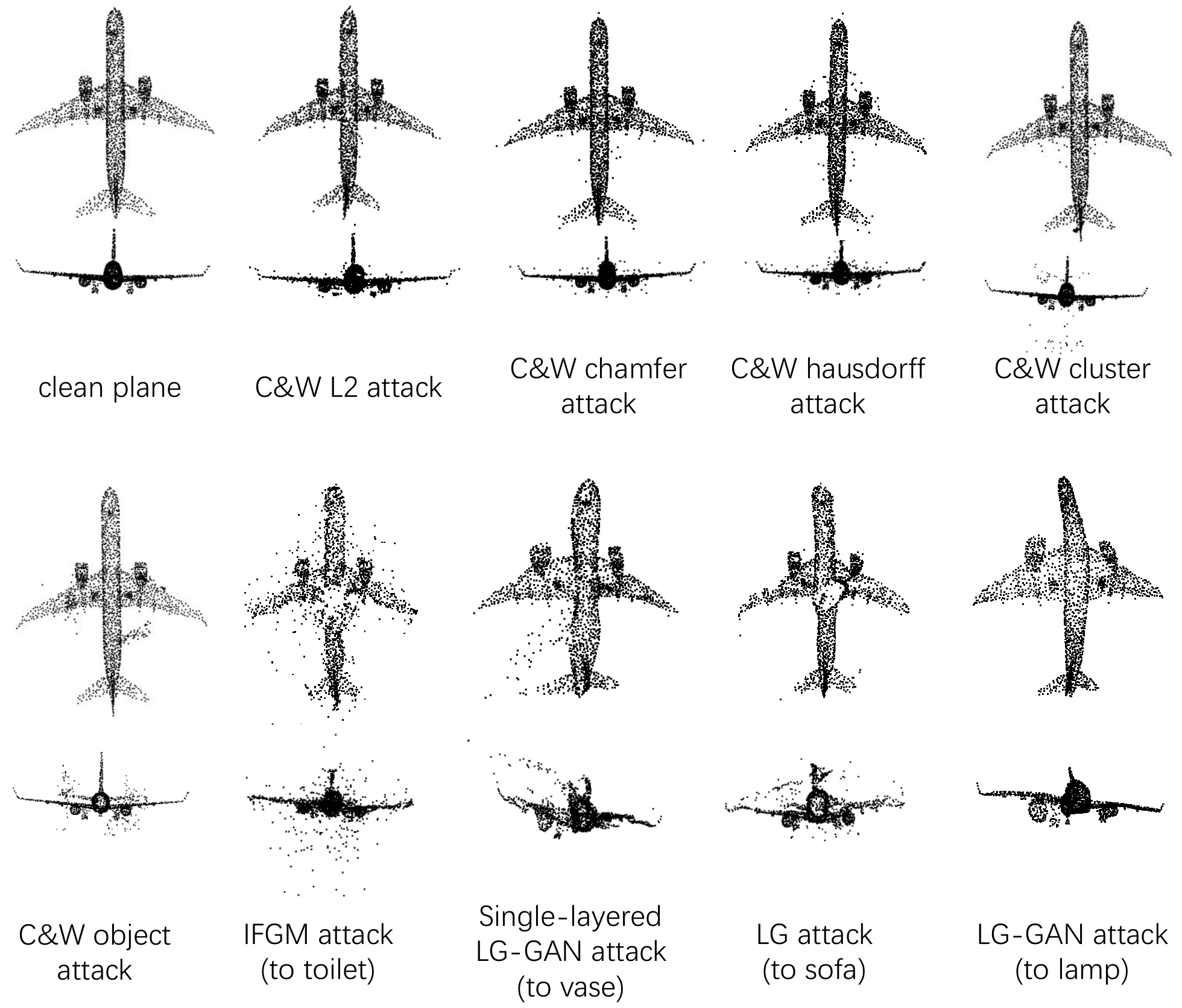}
}\\
{\centering\includegraphics[width=1.00\linewidth]{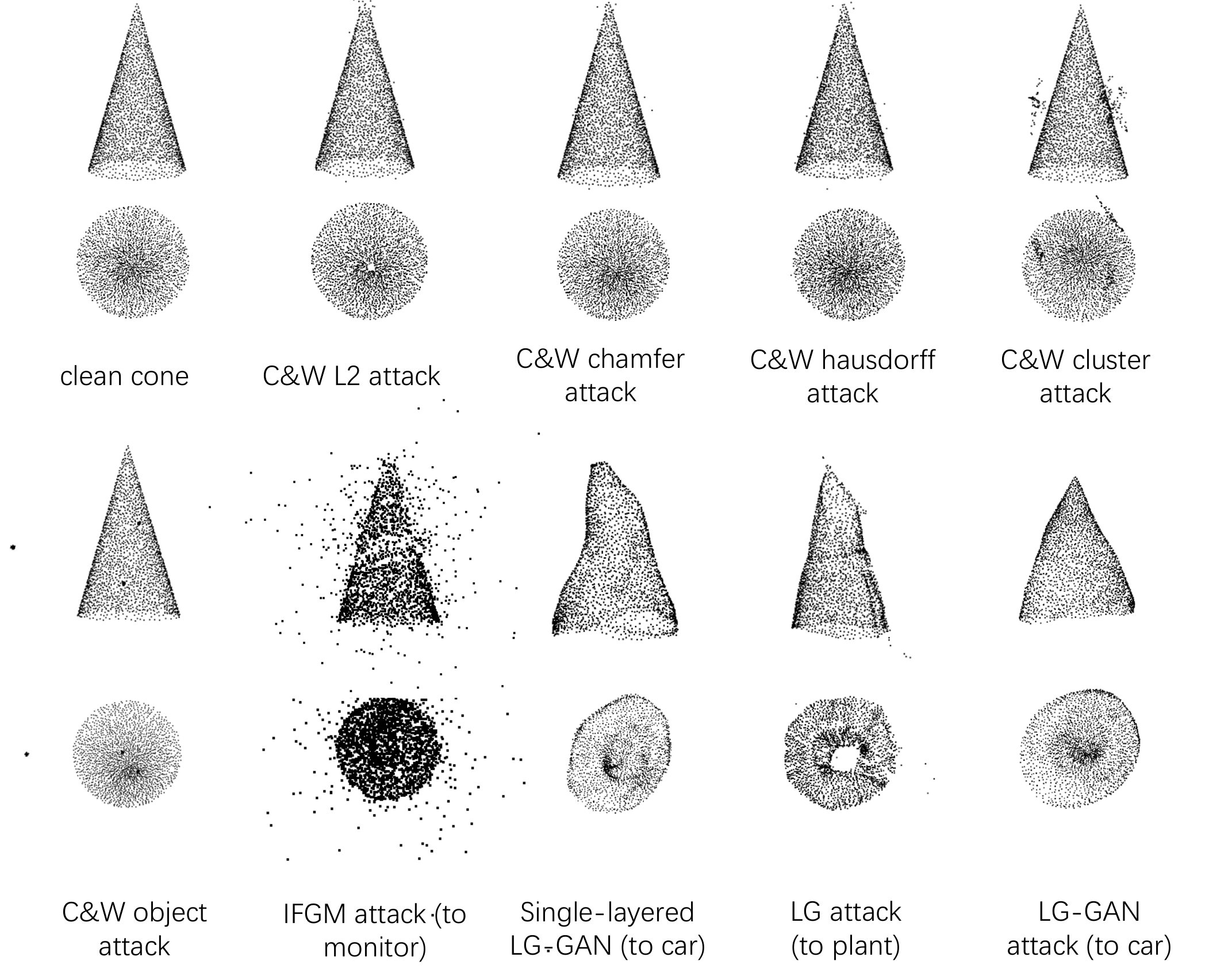}
}\\
\caption{\textbf{Qualitative results on two point clouds from ModelNet40, ``plane'' and ``cone''.} See Sec. \ref{qualitative} for details. Enlarge to see details. }
\label{fig:05}
\end{center}
\end{figure}

\subsection{Ablation Study}\label{ablation}

\vspace{0.5em}
\textbf{Multi-layered or single-layered concatenation?}
In this part, we explore the influence of multi-layered concatenation for reconstruction result. We retain the bottom label concatenation layer and train such single-layered LG-GAN. Since multi-layered LG-GAN concatenates features repeatedly along the convolution operation, it helps the point cloud feature and label feature bond together in a progressive way. In other words, it is easier for the multi-layered LG-GAN to attack successfully without sacrificing visual quality too much. As shown in Table~\ref{tab:01} and Fig.~\ref{fig:05}, under the same attack success rate, multi-layered LG-GAN performs better than single-layered one in terms of gray-box attacks and visual quality.

\vspace{0.5em}
\textbf{$\ell_2$ loss or Chamfer loss?}
Chamfer loss is a typical metric for constraining a desirable point cloud to the original sample, which is a looser loss compared with the point-to-point $\ell_2$ loss. Because the output of LG-GAN shares the same point number with the input sample, both  $\ell_2$ loss and Chamfer loss can serve as the reconstruction loss. We argue that since the generation network tries to deform the generated sample to fit the targeted class, it is more difficult to preserve the shape of the point cloud than tackling generation tasks such as surface reconstruction and point cloud upsampling. In the 3D space, Chamfer loss cannot restrain point drift to some extent, and thus visually produces isolated points, which are known as ``outliers''. 
Table~\ref{tab:01} shows that using $\ell_2$ loss performs better than Chamfer loss.

\vspace{0.5em}
\textbf{Effect of the GAN loss.}
Compared to LG-GAN, we remove the discriminator network from the proposed structure, denoted as LG. We argue that the GAN loss helps further constrain the manifold surface and increases surface smoothness, but at the same time bends the shape more. Another benefit of GAN loss is helping suppress the outliers and improving the attack ability on gray-box defense including SRS (+3.8\%) and DUP-Net~\cite{zhou2019dup} (+12.8\%). Both $\ell_2$ and Chamfer distances have slightly enlarged because despite point clouds being more compact, they are more distant from the original sample in the distance evaluation in the Cartesian coordinate system.


\vspace{0.5em}
\textbf{Visual quality \vs attack success rate.}
Weight factor $\alpha$ controls the balance between the visual quality of an adversarial point cloud and the attack success rate it has. Without the constraint of classification loss (when $\alpha$ is zero), the target of the network is to learn the shape representation of the input object. When $\alpha$ increases, more generated point clouds can successfully target-attack the recognition network, but visual quality gradually decreases. In Fig.~\ref{fig:alpha}, we show the average attack success rate and distortion evaluation (\ie $\ell_2$ and Chamfer distances) with varying $\alpha$. When $\alpha$ is set with 100, a superior attack success rate is achieved with 90\% or so,
and the Chamfer distance is around 0.007, which is slightly larger than C\&W with $\ell_2$ loss based attacks (0.006). Thus, arguably, LG-GAN is close to optimization-based attacks in most cases. Besides, in Fig.~\ref{fig:alpha_img} we visualize the generated point clouds with different $\alpha$. To reach 40\% attack success rates ($\alpha$ is 10), it is effortless for LG-GAN to reshape the input object a little bit to achieve the goal of attack with slight perturbation of points on the object surface. The object gradually bends out of shape when $\alpha$ increases, but barely produces scattered outliers compared with gradient-based and optimization-based methods.

The curvilinear trend also reveals that some categories are harder to be attacked targetedly than other categories. We average over 10 times of the class accuracy of adversarial point clouds with LG-GAN and its variances under $\alpha$ is 100, as shown in Fig.~\ref{fig:class_acc}(a). As can be seen, ``lamp'' and ``stairs'' are two categories that are hard to be operated to attack. Similar results can be obtained when attack success rates are low ($<50\%$). Furthermore, we visualize the objects from the two categories that fail to attack PointNet in Fig.~\ref{fig:class_acc}(b), and realize the attack ability of generating adversarial point clouds depends on the shape diversity among object classes. It is more difficult to deform ``lamp'' or ``stairs'' to other classes.
In short, shapes determine the difficulty level of attacks; it is unavoidable to sacrifice visual quality to increase attack ability.


\vspace{0.5em}
\textbf{Attack transferability.}
The results in Table.~\ref{tab:02} illustrate that all the attack schemes have limited transferability. On the whole, adversarial examples generated from PointNet~\cite{qi2017pointnet} have better attack success rate on DGCNN~\cite{wang2019dynamic} than PointNet++~\cite{qi2017pointnet++} which indicates that the structures of PointNet and DGCNN are more similar. Nevertheless, LG-GAN performs better than C\&W and IFGM methods because adversarial examples generated by LG-GAN have better fluid-like shapes than others, which are still able to fool other unknown networks.

\vspace{0.5em}
\textbf{Speed.}
Our proposed model is very efficient since it only accesses the target model in the inference stage only once. Compared to previous best methods (Table~\ref{tab:01}), our model is more than $1000\times$ times faster than C\&W and $7\times$ faster than IFGM in speed. Therefore, LG-GAN is more friendly to be utilized in real-time systems.

\subsection{Translation Attack}\label{translation}
We have designed an alternative attack based on geometric translation. It is observed that the centroids of generated adversarial point clouds are not on the origin of the Cartesian coordinate system compared with the original point clouds. By normalizing to center, attack success rate of LG-GAN has dropped slightly ($98.2\%\rightarrow 82.6\%$), still better than IFGM ($73.0\%\rightarrow 72.8\%$) and C\&W with $\ell_2$ loss ($100\%\rightarrow 0\%$).
We proceed with a second analysis by translating a whole point cloud to random direction and find that the networks are fragile to monolithic translation. More details are given in the supplementary material.


\subsection{Qualitative Results}\label{qualitative}
Fig.~\ref{fig:05} shows two representative examples of adversarial point clouds (``plane'' and ``cone'') using C\&W~\cite{xiang2019generating}, IFGM~\cite{liu2019extending,yang2019adversarial} and LG-GAN ($\alpha$ is 40) schemes from ModelNet40 dataset. As can be seen, C\&W with $\ell_2$, Chamfer and Hausdorff losses have the least perturbations among the three schemes, but create outliers far away from the object surface relatively. For C\&W with cluster and object losses, they only add points and the original point clouds are not modified. For IFGM, plenty of points are moved to outside or inside of the object, which makes the object more obscure to observe. For LG, without GAN loss, the shapes of point clouds are preferable and easy to recognize. For LG-GAN, due to the constraint of GAN loss, the density of points on the surface has been changed but the points still adjoin the manifold, which is more difficult to recognize. However, as stated above, the advantage of LG-GAN lies in that it can attack outlier-removal based defences with a high success ratio. In a word, there is still a long way to design a nice and distortionless adversarial point cloud that can totally deceive human eyes.


\section{Conclusion}

In this work we have introduced LG-GAN: a novel label guided adversarial network for arbitrary-target point cloud attack. By feeding the original point clouds and target attack label into LG-GAN, it can learn how to deform the point cloud with minimal perturbations to mislead the recognition network into the specific label only with a single forward pass. In details, LG-GAN first leverages one multi-branch adversarial network to extract  hierarchical features of the input point clouds, then utilizes a label encoder to incorporate the specified label information into multiple intermediate features. Finally, the transformed features will be fed into the coordinate reconstruction decoder to generate the target adversarial sample. Experiments show that the proposed LG-GAN can support more flexible targeted attack on the fly while guaranteeing good attack performance and higher efficiency simultaneously.


\textbf{Acknowledgement} This work was supported in part by the Natural Science Foundation of China under Grant U1636201 and 61572452, and by Anhui Initiative in Quantum Information Technologies under Grant AHY150400. Gang Hua was supported partly by National Key R\&D Program of China Grant 2018AAA0101400 and NSFC Grant 61629301.

{\small
\bibliographystyle{ieee_fullname}
\bibliography{pcGenerate}
}

\end{document}


\title{LG-GAN: Label Guided Adversarial Network for Flexible Targeted Attack of Point Cloud-based Deep Networks Supplementary Material}

\newcommand*{\affaddr}[1]{#1} 
\newcommand*{\affmark}[1][*]{\textsuperscript{#1}}
\newcommand*{\email}[1]{\texttt{#1}}

\renewcommand\thesection{\Alph{section}}
\renewcommand\thesubsection{\thesection.\alph{subsection}}

\author{Hang Zhou$^{1}$\thanks{Equal contribution, $\dagger$ Corresponding author}, Dongdong Chen$^{2,*}$, Jing Liao$^{3}$,  Kejiang Chen$^{1}$ , Xiaoyi Dong$^{1}$, \\ Kunlin Liu$^{1}$, Weiming Zhang$^{1,\dagger}$, Gang Hua$^{4}$, Nenghai Yu$^{1}$\\
$^{1}$University of Science and Technology of China, \quad $^{2}$Microsoft Research \\
$^{3}$City University of Hong Kong, $^{4}$Wormpex AI Research
\\
{\tt\small\{zh2991,chenkj,dlight,lkl6949\}@mail.ustc.edu.cn}; \tt\small\email{cddlyf@gmail.com};\\ \tt\small\email{jingliao@cityu.edu.hk}; {\tt\small\email{\{zhangwm, ynh\}@ustc.edu.cn}};
\tt\small\email{ganghua@gmail.com}
}

\maketitle
\ifcvprfinal\thispagestyle{empty}\fi

\section{Overview}

This document provides additional quantitative results, technical details and more qualitative test examples to the
main paper.

In Sec.~\ref{B} we provide more details on neural network architectures and training parameters.
In Sec.~\ref{C} we extend the attack performance on PointNet++~\cite{qi2017pointnet++} with ModelNet40. In Sec.~\ref{D} we extend the attack performance on PointNet~\cite{charles2017pointnet} with ShapeNet. 
In Sec.~\ref{E} we design a new metric for evaluating adversarial point cloud effect. 
In Sec.~\ref{F} we put forward a construction for adaptively designing adversarial point clouds with different deformation degrees. 
In Sec.~\ref{G} we design an alternative attack based on geometric translation, and in Sec.~\ref{H} we give more visualization results.

\section{Details of Network Architectures}\label{B}

The details of our network architecture are described as follows:

In the hierarchical feature learning component of the point cloud encoder, we utilize 4 levels to extract local features. Following the notations in PointNet++~\cite{qi2017pointnet++}, we utilize $(m, r, [l_1, ..., l_d])$ to represent a level with $m$ local regions of ball radius $r$ with 32 adjacent points, and $[l_1, ..., l_d]$ represents the $d$th FC layers with width $l_i(i=1, ..., d)$. Therefore, the parameters we use are $(N, 0.05, [32, 32, 64])$, $(\frac{N}{2}, 0.1, [64, 64, 128])$, $(\frac{N}{4}, 0.2, [128, 128, 256])$ and $(\frac{N}{8}, 0.3, [256, 256, 512])$.

In the decoder side, we utilize interpolation to restore the feature of each level and use a convolution to reduce the restored feature to 64 dimensions. We then utilize aggregation to merge multiple layers extracted from different scales together. We utilize three FC layers which in between concatenated with label features, and the output feature channel numbers are 256, 128 and 64, respectively. Finally, we utilize a FC layer with 3 output channels to reconstruct the final coordinates. Note that the convolution layers and FC layers are followed by the ReLU activation layers, except for the last coordinate reconstruction layer.

\begin{figure*}
\begin{center}
{\centering\includegraphics[width=0.65\linewidth]{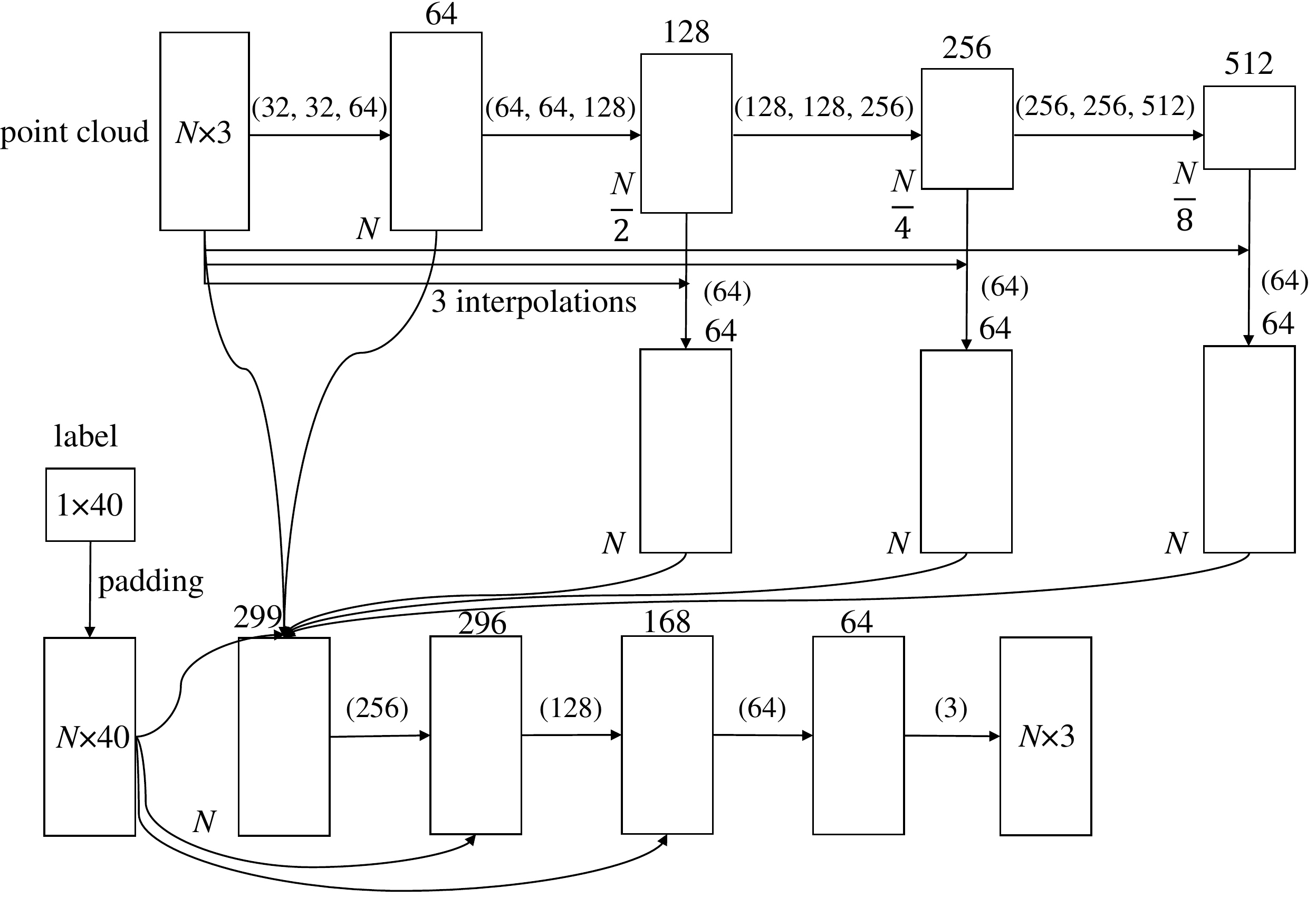}
}\\
\caption{\textbf{The generator part of the network architecture of LG-GAN.} }
\label{fig:01}
\end{center}
\end{figure*}

The details of the baseline architectures are illustrated in Fig.~\ref{fig:01}.

\begin{table}
\footnotesize
\begin{center}
\begin{tabular}{c|c|c|c}
\hline\toprule[0.4pt]
 & Target~\cite{qi2017pointnet++} & \tabincell{c}{Defense\\(SRS)~\cite{zhou2019dup}} & \tabincell{c}{Defense\\ (DUP-Net)~\cite{zhou2019dup}} \\
\midrule
C\&W $+$ $\ell_2$~\cite{xiang2019generating}    & \textbf{100} & 0 & 0 \\
C\&W $+$ Hausdorff~\cite{xiang2019generating}   & \textbf{100} & 0 & 0 \\
C\&W $+$ Chamfer~\cite{xiang2019generating}     & \textbf{100} & 0 & 0 \\
C\&W $+$ 3 clusters~\cite{xiang2019generating}  & 93.3 & 3.5 & 0 \\
C\&W $+$ 3 objects~\cite{xiang2019generating}   & 97.3 & 0.4 & 0 \\
FGSM~\cite{liu2019extending,yang2019adversarial}& 0.1 & 0 & 0 \\
IFGM~\cite{liu2019extending,yang2019adversarial}& 4.9 & 0 & 0 \\
\midrule
LG-GAN (ours)                            & 50.4 & \textbf{40.8} & \textbf{45.1} \\
\hline\toprule[0.4pt]
\end{tabular}
\end{center}
\caption{\textbf{Attack success rate (\%, second to fourth column) on attacking PointNet++~\cite{qi2017pointnet++} from ModelNet40.} ``Target'' stands for white-box attacks. The hyper-parameter setting of two gray-box attacks is: for the simple random sampling (SRS) defense model, percentage of random dropped points is 60\%$\sim$90\%; for DUP-Net defense model, $k=50$ and $\alpha=0.9$ from~\cite{zhou2019dup}. The default LG-GAN (ours) consists of multi-layered label embedding, $\ell_2$ loss and GAN loss. }
\label{tab:01}
\end{table}

\section{Comparing with State-of-the-art Methods Generated from PointNet++}\label{C}
\vspace{0.5em}
Results of attacking PointNet++~\cite{qi2017pointnet++} are summarized in Table~\ref{tab:01}. FGSM and IFGM have 0.1\% and 4.0\% attack success rates respectively, which are much lower than attacking PointNet~\cite{charles2017pointnet}. LG-GAN outperforms IFGM methods by at least 45.1\% of attack success rates. C\&W based methods still can reach near 100\% attack success rates when attacking PointNet++, but LG-GAN can only reach 50\% attack success rate, which can be attributed to the fact that PointNet++ has more complicated network structures which is more difficult to attack. In terms of gray-box attacks, LG-GAN still has better attack ability, with 40.8\% and 45.1\% attack success rates on simple random sampling (SRS) and DUP-Net~\cite{zhou2019dup} defense model, respectively; while for optimization-based C\&W methods and gradient-based FGSM and IFGM, they all fail to attack. It should be noted that LG-GAN is still the fastest attack method among them.

\section{Comparing with State-of-the-art Methods Generated from PointNet under ShapeNet}\label{D}
The results are summarized in Table~\ref{tab:02}. Similar to the results on ModelNet40, LG-GAN has more than 90\% white-box attack success rates, and performs better than existing attacks in terms of attack success rates on defense models. 

\begin{table}
\footnotesize
\begin{center}
\begin{tabular}{c|c|c|c}
\hline\toprule[0.4pt]
 & Target~\cite{charles2017pointnet} & \tabincell{c}{Defense\\(SRS)~\cite{zhou2019dup}} & \tabincell{c}{Defense\\ (DUP-Net)~\cite{zhou2019dup}} \\
\midrule
C\&W $+$ $\ell_2$~\cite{xiang2019generating}    & 100  & 0.6  & 0.1  \\
C\&W $+$ Hausdorff~\cite{xiang2019generating}   & 100  & 0.4  & 0.1  \\
C\&W $+$ Chamfer~\cite{xiang2019generating}     & 100  & 0.5  & 0.1  \\
C\&W $+$ 3 clusters~\cite{xiang2019generating}  & 100  & 0.5  & 0.1  \\
C\&W $+$ 3 objects~\cite{xiang2019generating}   & 100  & 0.5  & 0.1  \\
FGSM~\cite{liu2019extending,yang2019adversarial}& 0    & 0    & 0    \\
IFGM~\cite{liu2019extending,yang2019adversarial}& 67.5 & 2.6  & 2.3  \\
\midrule
LG-GAN ($\alpha=1000$)                          & 98.6 & 98.4 & \textbf{62.1} \\
LG-GAN ($\alpha=5000$)                          & 93.9 & 92.9 & \textbf{58.9} \\
\hline\toprule[0.4pt]
\end{tabular}
\end{center}
\caption{\textbf{Attack success rate (\%, second to fourth column), distance (fifth-sixth column) between original sample and adversarial sample (meter per object) and generating time (second per object) on attacking PointNet from ShapeNet.} ``Target'' stands for white-box attacks. The hyper-parameter setting of two gray-box attacks is: for the simple random sampling (SRS) defense model, percentage of random dropped points is 60\%; for DUP-Net defense model, $k=50$ and $\alpha=0.9$ from~\cite{zhou2019dup}. The default LG-GAN (ours) consists of multi-layered label embedding, $\ell_2$ loss and GAN loss. }
\label{tab:02}
\end{table}

\section{Perturbation Metric Comparison}\label{E}

\begin{table}
\footnotesize
\begin{center}
\begin{tabular}{c|ccccc}
\hline\toprule[0.4pt]
 & \tabincell{c}{Clean\\data} & IFGM & CW+$\ell_2$ & \tabincell{c}{CW+\\Chamfer} & \tabincell{c}{LG-GAN \\($\alpha=1000$)} \\
\midrule
$\ell_2$ & --- & 0.31 & 0.01 &  0.1 & 0.35 \\
Kurtosis & 5.3 & 48.3 & 48.9 & 72.4 & \textbf{44.1} \\
\hline\toprule[0.4pt]
\end{tabular}
\end{center}
\caption{\textbf{Perturbation metric comparison among CW, IFGM and LG-GAN.}  }
\label{tab:03}
\end{table}

We design a kurtosis based perturbation metric to evaluate adversarial effect more accurately. 
Although C\&W attacks have smaller $\ell_2$ distances than LG-GAN's attack, they will create distinct visual outliers. To effectively measure visual distortion, we have designed a point-density based evaluation function $K(\mathcal{P})$, \ie the kurtosis (the sharpness of the peak of a frequency-distribution curve) of the sorted set of nearest distances of all the points. Specifically,
\begin{equation}
\label{eqn:01}
K(\mathcal{P})=kurtosis(sort(nearest\_{\ell_2}(\mathcal{P}))).
\end{equation}
We have verified that $K_{IFGM}(\mathcal{P})>K_{C\&W}(\mathcal{P})>K_{LG-GAN}(\mathcal{P})>K_{ORIG}(\mathcal{P})$. The numerical results is given in Table~\ref{tab:03}.

\section{Adaptive Deformation Degree}\label{F}
In general, training multiple models with different $\alpha$ is one most straightforward way to tune the deformation degree. But the proposed framework is very general and can support it in a smarter way. Specifically, we can input $\alpha$ as the extra condition into the encoder $\mathbf{E}_l$ and train one single LG-GAN with randomly sampled
$\alpha$. 

\section{Translation Attack}\label{G}

\begin{table}
\footnotesize
\begin{center}
\begin{tabular}{l|c|c|c}
\hline\toprule[0.4pt]
$\epsilon$ (meter) & PointNet~\cite{qi2017pointnet} & PointNet++~\cite{qi2017pointnet++} & DGCNN~\cite{wang2019dynamic} \\
\midrule
 0    & 88.6 & 89.5 & 87.9 \\
0.01  & 88.5 & 89.5 & 87.8 \\
0.1   & 77.2 & 89.6 & 87.2 \\
0.5   & 13.2 & 89.7 & 57.9 \\
1     &  3.5 & 86.4 & 14.4 \\
2     &  1.7 & 61.9 &  4.1 \\
10    &  2.0 &  6.0 &  2.6 \\
\hline\toprule[0.4pt]
\end{tabular}
\end{center}
\caption{\textbf{Detection accuracy (\%) of point-cloud translation attacks} on deep networks~\cite{qi2017pointnet,qi2017pointnet++,wang2019dynamic} of ModelNet40. $\epsilon$ is the maximum stride size of translating one whole point cloud along X-axis, Y-axis and Z-axis.  }
\label{tab:04}
\end{table}

We have designed an alternative attack based on geometric translation. It is observed that the centroids of generated adversarial point clouds are not on the origin of the Cartesian coordinate system compared with the original point clouds. 
We proceed with a second analysis by translating a whole point cloud to random direction and find that the networks are fragile to monolithic translation. We move the normal point cloud along XYZ directions with different stride size following the uniform distribution between 0 and maximum stride size $\epsilon$, where $\epsilon$ is 0, 0.01, 0.1, 0.5, 1, 2 and 10. These are untargeted attacks, which deviate the network prediction from the original label. The results are summarized in Table~\ref{tab:04}. Larger point-cloud offset tends to deteriorate the classification result even more. It is shown that PointNet++ is more robust against translation attack compared with PointNet and DGCNN, but still fails to defend against adversarial point clouds with a large offset ($\epsilon>1$).

\section{More Visualizations}\label{H}
\begin{figure*}
\begin{center}
{\centering\includegraphics[width=1.00\linewidth]{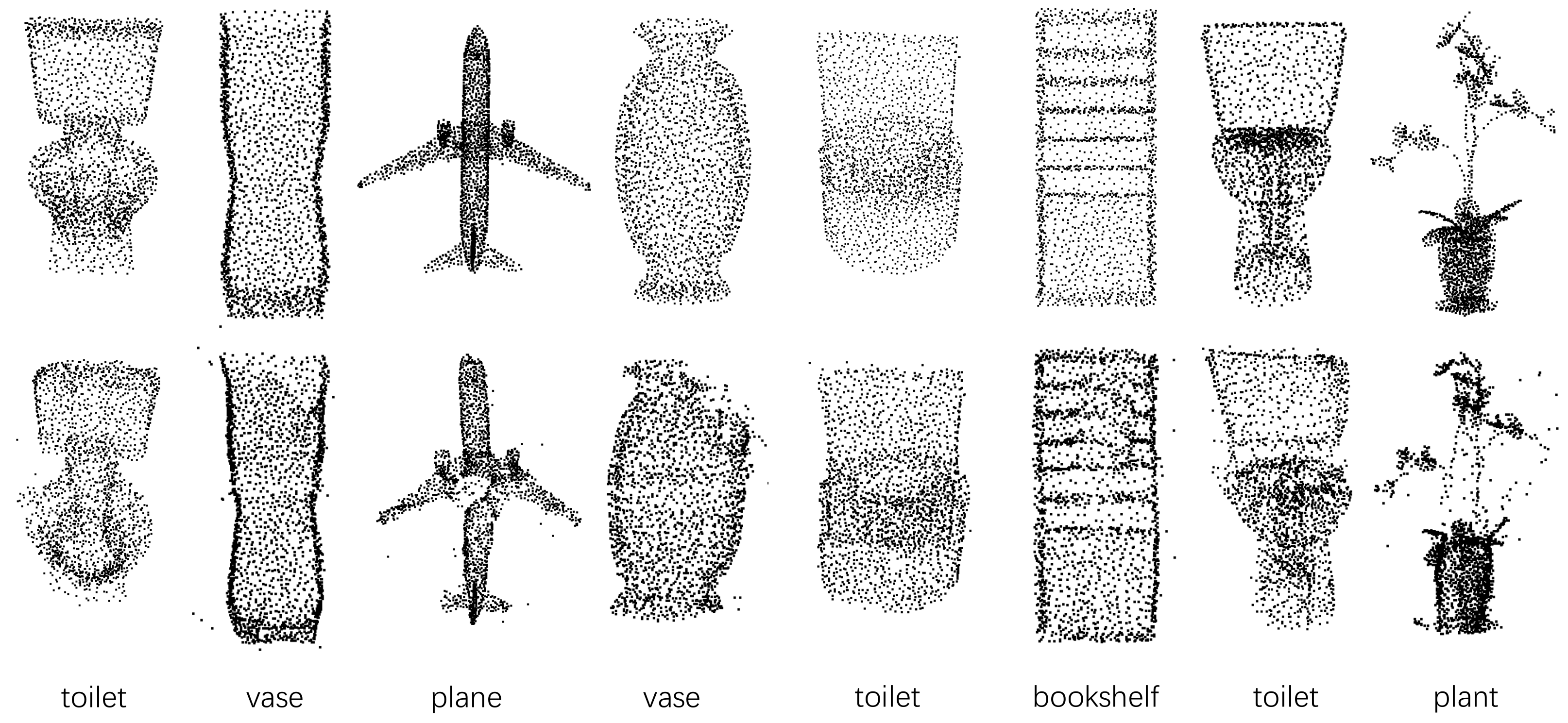}
}\\
{\centering\includegraphics[width=1.00\linewidth]{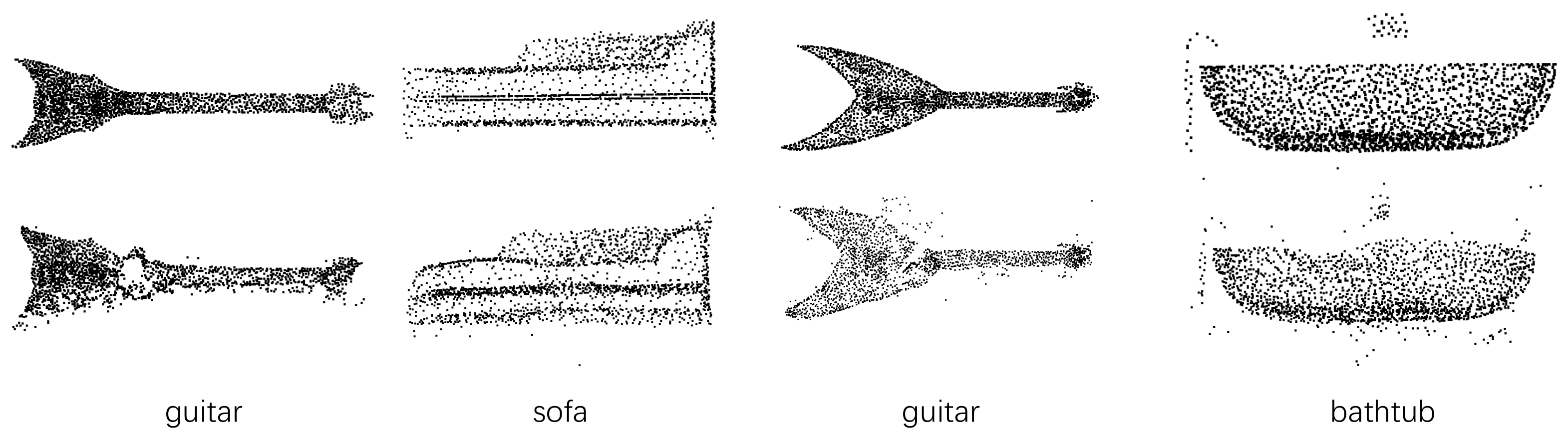}
}\\
{\centering\includegraphics[width=1.00\linewidth]{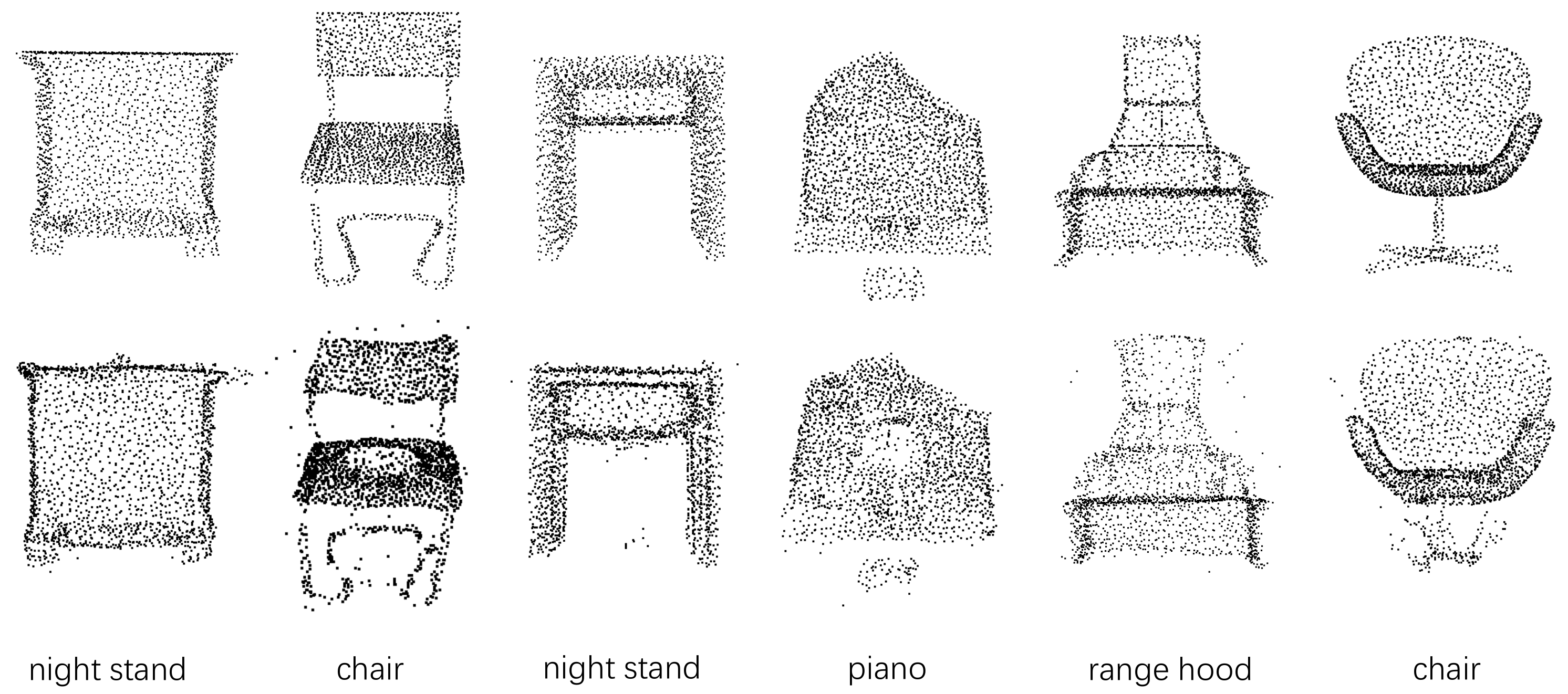}
}\\
\caption{\textbf{Qualitative results of targeted attacks on ModelNet40.} We attack PointNet to random arbitrary labels (except for the original label). The odd-numbered lines are the original point clouds and the even-numbered lines are the corresponding adversarial point clouds. Enlarge to see details. }
\label{fig:02}
\end{center}
\end{figure*}

We present more results of adversarial point clouds by attacking PointNet~\cite{charles2017pointnet} compared with the original samples on ModelNet40 in Fig.~\ref{fig:02}.

{\small
\bibliographystyle{ieee_fullname}
\bibliography{supplement}
}